\documentclass[11pt]{article}
\usepackage[final]{acl}
\usepackage{times}
\usepackage{latexsym}
\usepackage[T1]{fontenc}
\usepackage[utf8]{inputenc}
\usepackage{microtype}
\usepackage{booktabs}
\usepackage{amsmath}
\usepackage{amssymb}
\usepackage{graphicx}
\usepackage{xcolor}
\usepackage{tikz}
\usetikzlibrary{arrows.meta,positioning}
\definecolor{stubfill}{RGB}{220,230,242}
\definecolor{livefill}{RGB}{255,236,200}
\DeclareRobustCommand{\code}[1]{\texttt{\detokenize{#1}}}
\newcommand{\Nincidents}{6}
\newcommand{\liveCalls}{12}
\newcommand{\faultCorrect}{6}
\newcommand{\determReps}{20}
\newcommand{\benignSurvived}{30}
\newcommand{\boundarySites}{18}
\newcommand{\replayCostK}{0.00}
\newcommand{\replaySuiteMs}{3.5}
\newcommand{\cutpointSuiteMs}{3.6}
\newcommand{\recOverheadUs}{23}
\newcommand{\recOverheadPct}{0.008}
\newcommand{\modelLatencyMs}{300}
\newcommand{\storeKB}{1.44}
\newcommand{\mutantsTotal}{192}
\newcommand{\mutantsKilledCP}{51}
\newcommand{\mutantsKilledFS}{0}
\newcommand{\mutantsSurvived}{141}
\newcommand{\mutantsUnkillable}{110}
\newcommand{\mutantsIgnoredField}{31}
\newcommand{\mutantsKillable}{82}
\newcommand{\mutantsKillableScore}{62}
\newcommand{\fullstubSpec}{0}
\newcommand{\liveModel}{Qwen3.5~4B}
\newcommand{\liveN}{50}
\newcommand{\liveOnMean}{3{,}136}
\newcommand{\liveOnStd}{379}
\newcommand{\liveOffMean}{3{,}045}
\newcommand{\liveOffStd}{387}
\newcommand{\liveDiff}{$+91$}
\newcommand{\liveCIlo}{$-59$}
\newcommand{\liveCIhi}{$+241$}
\title{Chronicle: Cut-Point Replay for Regression Testing of LLM Agents}

\author{Tisha Chawla\thanks{Equal contribution.} \quad
  Susheem Koul\footnotemark[1] \\
  Microsoft \\
  \texttt{tisha.chawla2020@vitalum.ac.in} \\
  \texttt{f2015347p@alumni.bits-pilani.ac.in}}

\begin{document}
\maketitle
\begin{abstract}
Large language model responses are non-deterministic, so failures in LLM agents
are hard to reproduce: a failure depends on inference that is not bitwise
reproducible \citep{atil-etal-2025-non}, on tools that read changing state, and
on a multi-step trajectory that a re-run rarely repeats. Record-and-replay makes
a run reproducible, but existing agent tooling records runs only to trace or
score them, not to test a code change against them. We present Chronicle, which
records an agent run at its non-deterministic boundaries as immutable envelopes
and replays it from the record. Its central operation, \emph{cut-point replay},
serves a chosen subset of boundaries from the record and executes the
complementary subset live with new code, turning a recorded incident into a
regression test that runs in continuous integration. On a benchmark of
\Nincidents{} recorded failures with simulated model boundaries, recording adds
\recOverheadUs\,\textmu s per crossing (\recOverheadPct\% of an assumed
\modelLatencyMs~ms model call), full replay issues zero model calls and is
bit-stable across \determReps{} repetitions, and cut-point tests fail on faulty
code and pass on guarded and benign changes for all \Nincidents{} incidents. In
a mutation study of the guarded tools, cut-point tests catch every mutant that
lets the recorded unsafe action through, while a baseline that stubs every
boundary, using the same assertion, catches none. Chronicle and the benchmark
are publicly available at \url{https://github.com/theagentplane/chronicle}.
\end{abstract}

\section{Introduction}
Large language model (LLM) agents take consequential actions through tool calls,
such as issuing refunds, submitting trades, or deleting files
\citep{schick2023toolformer, yao2023react}. When such an action is wrong,
because the surrounding code misses an edge case or the task is underspecified,
correcting it requires first reproducing the failure. This is often the hardest
step: a failure is seldom a single wrong output but a \emph{trajectory} of model
calls, tool calls, and routing decisions \citep{cemri2025why}. Re-executing the
agent rarely reproduces that trajectory. LLM inference is not bitwise
reproducible even at temperature zero \citep{atil-etal-2025-non}; tools read
external state that has since changed \citep{yao2024taubench}; and retries and
routing change how many
times a step runs \citep{cemri2025why}. The result is \emph{flakiness}:
identical code passes or fails depending on nondeterministic conditions and
execution order \citep{lam2020owl, hashemi2025detecting}, so re-execution alone
cannot turn an incident into a repeatable test.
Existing agent infrastructure observes runs but does not make them testable:
tracing records what happened \citep{arize2026phoenix} and evaluation frameworks
score whether an output is acceptable \citep{promptfoo2026, zheng2023judging},
but neither lets a developer change one component of a recorded run and check
whether the change fixes the failure while everything else is held fixed. We
present Chronicle, which supplies exactly that operation. This paper contributes
(i)~\emph{cut-point replay}, which runs new code at a chosen subset of boundaries
while serving the rest from the record, producing a committed regression test
from the incident, with each stubbed boundary guarded by a per-name call-count
check; and (ii)~an implementation and a benchmark of \Nincidents{} recorded
failures on which full replay makes no model call and reproduces each run
identically across \determReps{} repetitions, while cut-point tests flag every
unguarded incident, accept its guarded fix and benign edits, and kill mutants of
the fix that a stub-every-boundary baseline cannot.

\section{Related Work}
\label{sec:related}
Chronicle relates to three lines of work and differs from each in the purpose for
which replay is used.

\paragraph{Record and replay for agents.} Deterministic record-replay is long
established for debugging general programs, where an entire execution is
captured and faithfully reproduced \citep{ocallahan2017rr}. Chronicle instead
replays at semantic boundaries and runs new code at chosen ones, so replay tests
a fix rather than reproduces a run. Prior systems record runs to reuse successful
agent behavior \citep{feng2025experience} or to resume a durable workflow after a
failure \citep{temporal2026}; Temporal's replay tests also re-run changed
workflow code against a recorded event history, and Chronicle adapts this
selective-replay idea to agent boundaries, with indexed crossings and a freely
chosen live set. Graph frameworks checkpoint state to permit re-entry at a
previously executed node \citep{langgraph2026}. These systems act \emph{during}
a run, to resume or steer it in flight; a workflow resumed after a
human-in-the-loop pause continues forward and does not re-issue the model calls
it already completed. Chronicle instead operates \emph{after} a run has finished:
it replays a recorded trace, serving stubbed crossings from the record and
running the live subset with new code, so a candidate fix is evaluated against
the past incident rather than by steering a live run.

\paragraph{Testing of agents.} Recent work adapts software-testing techniques
such as traces, mocks, and assertions to agents \citep{kohl2026automated}. A
documented limitation of heavy mocking is that mock-heavy tests can be less
effective at validating real interactions \citep{hora2026coding}. Chronicle
replaces hand-written mocks with envelopes drawn from a recorded run and lets any
subset of boundaries run live, so a tool gate or router change is exercised
against the recorded trajectory rather than against invented stubs.

\paragraph{Failure attribution.} A related line of work intervenes in a recorded
run (rewinding, editing, or otherwise modifying a crossing) to attribute or
localize a failure \citep{shah2026causal, lin2026reflect, ma2025dover}. Those
methods characterize the cause of a past failure by observing a modified
crossing, whereas Chronicle executes new code at the crossing and records the
outcome as a regression test. Other tools detect and diagnose faults from traces
without cut-point execution of new agent code
\citep{balusu2026agenttelemetry, ou2025agentdiagnose, deshpande2025trail,
kang2026knowledge}, and LLM-based judges assess semantic quality subject to
documented limitations in agreement and bias \citep{norman2026reliability};
Chronicle ships an advisory LLM-as-judge for non-structural properties but does
not evaluate judge reliability here.

The agents considered in this work employ standard reason-and-act loops with
tool use \citep{yao2023react, schick2023toolformer}. In contrast to hand-authored
agent benchmarks and evaluation environments
\citep{yao2024taubench, debenedetti2024agentdojo, zhou2024webarena}, the
scenarios in our benchmark are derived from recorded failures rather than
constructed to evaluate capability or attack surface.

\section{Chronicle}
\label{sec:chronicle}
Chronicle records a run, replays it from the record, and tests the replay with
assertions. Its unit is the \emph{boundary}: a point where the agent calls the
model, calls a tool, or makes a routing decision, which is exactly where a rerun
can diverge. Chronicle adapts selective replay to agents: any subset of
boundaries can run live while the rest are served from a real recorded run
rather than from hand-written mocks, so a recorded incident becomes a
deterministic test of a chosen change.

\subsection{Record}
Developers mark boundaries with a one-line annotation (Figure~\ref{fig:api}); a
model client or a set of LangGraph nodes can be instrumented with a single call.
Each execution of a boundary, a \emph{crossing}, is saved as an immutable
\emph{envelope} holding its input, its output, and the metadata needed to detect
drift, such as the model version and sampling parameters. Crossings are
addressed by boundary name and occurrence, so a boundary crossed three times in
a loop yields \code{agent[1]}, \code{agent[2]}, and \code{agent[3]}
(Figure~\ref{fig:indexing}). Recording is transparent: it changes no return
value or exception, redacts secrets and volatile fields before storage, and
emits standard OpenTelemetry spans. An envelope stores a boundary's input and
output, not the work inside it, so replaying it is faithful as long as its
output depends only on that recorded input; a boundary that reads hidden state,
such as a clock or a database, is the exception.

\subsection{Replay}
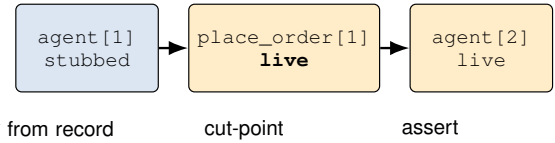
\begin{figure}[t]
\centering
\resizebox{\columnwidth}{!}{%
\begin{tikzpicture}[
  font=\scriptsize\ttfamily,
  box/.style={draw, rounded corners=2pt, align=center, minimum width=1.7cm,
              minimum height=1.05cm, inner sep=3pt},
  stub/.style={box, fill=stubfill},
  live/.style={box, fill=livefill},
  arr/.style={-{Latex}, thick},
  note/.style={font=\scriptsize\sffamily, align=left},
]
\node[stub] (a1) {agent[1]\\stubbed};
\node[live, right=0.35cm of a1] (t1) {place\_order[1]\\\textbf{live}};
\node[live, right=0.35cm of t1] (a2) {agent[2]\\live};
\draw[arr] (a1) -- (t1);
\draw[arr] (t1) -- (a2);
\node[note, below=0.2cm of a1, text width=1.9cm] {from record};
\node[note, below=0.2cm of t1, text width=1.9cm] {cut-point};
\node[note, below=0.2cm of a2, text width=1.9cm] {assert};
\end{tikzpicture}
}
\caption{Cut-point replay (trade-notional fixture): \texttt{agent[1]} stubbed;
\texttt{place\_order[1]} and \texttt{agent[2]} live; stubbed crossings return
their recorded outputs.}
\label{fig:cutpoint}
\end{figure}

In \emph{full replay}, every boundary returns its recorded output, so the run is
reproduced exactly with no model call (Figure~\ref{fig:fullreplay}). In
\emph{cut-point replay}, a chosen subset of crossings run live with new code
while the rest are served from the record (Figure~\ref{fig:cutpoint}); the live
subset is arbitrary (e.g., run a tool gate live while stubbing the model;
Figure~\ref{fig:multilive} in Appendix~\ref{app:figures}). Full replay fixes
every output and so tests the deterministic glue code between boundaries against
real recorded inputs; cut-point replay tests a change together with its
consequences, reproducing the stubbed lead-up without re-running it and
executing the live subset forward. Chronicle serves the $k$th crossing of a
boundary its $k$th recorded envelope, so if a stubbed boundary is crossed more
or fewer times than recorded (an extra loop, a dropped retry) lookup fails and
replay raises. This per-name count check does not detect a reordering that
preserves each name's count. An order-sensitive digest over the stubbed
crossings would close this gap cheaply; it is not part of the release evaluated
here. The guarantee is conditional on coverage: an unmarked non-deterministic
call runs live and escapes the check.

\subsection{Test}
A structural assertion checks what the agent did (which tool, which arguments,
whether a guarded action was refused); on full replay it is deterministic and
needs no model call, so a recorded incident and one assertion become a CI
regression test at no cost, and on cut-point replay the same assertion checks
the fix on the live subset. For non-structural properties (faithfulness, safety)
Chronicle ships an advisory LLM-as-judge \citep{zheng2023judging}; its
reliability is not evaluated here \citep{norman2026reliability}.

\begin{figure}[t]
\centering
\begin{minipage}{\columnwidth}
\footnotesize
\begin{verbatim}
@boundary("place_order", kind="tool")
def place_order(symbol, qty): ...

plan = (ReplayPlan()
        .stub("agent", 1)      # from record
        .live("place_order", 1) # cut-point
        .live("agent", 2))
assert session.captured_result(
    "place_order", 1)["blocked"]
\end{verbatim}
\end{minipage}
\caption{Recording annotation and a cut-point plan: the same \texttt{@boundary}
records in production and serves its recorded output in replay; any subset of
crossings may run live.}
\label{fig:api}
\end{figure}

\section{The Incident Benchmark}
\label{sec:benchmark}
We release a benchmark of \Nincidents{} recorded incidents, each a small agent
(model, then tool, then model) in which an unguarded tool produces an unsafe
result and a guarded version corrects it: a refund sized to an order identifier,
a currency mismatch, a notional-versus-quantity trade error, an over-broad email
audience, a payout-account injection, and a production file deletion. Each ships
with the recorded trace, the unguarded and guarded tools, and a cut-point test.
Each incident instantiates a documented failure mode from the multi-agent failure
taxonomy of \citet{cemri2025why}, chiefly its task-verification category (missing
or weak checks before acting), rather than an arbitrary bug
(Table~\ref{tab:incidents}). All
\Nincidents{} agents use deterministic simulated boundaries, so the harness makes
no provider API calls and runs in continuous integration; fixtures are
versioned.

\begin{table}[t]
\centering
\small
\setlength{\tabcolsep}{4pt}
\begin{tabular}{@{}lll@{}}
\toprule
Incident & Unsafe result & Guard \\
\midrule
Refund   & refund sized to order id & flat cap \\
Invoice  & wrong currency sent      & currency check \\
Trade    & notional read as shares  & notional cap \\
Email    & audience too broad       & recipient allowlist \\
Payout   & account substituted      & account check \\
Deletion & production file removed  & delete gate \\
\bottomrule
\end{tabular}
\caption{The \Nincidents{} recorded incidents. Each unguarded tool produces the
unsafe result; the guarded version applies the guard, and the cut-point test
asserts that the guard fired.}
\label{tab:incidents}
\end{table}

\paragraph{Why cut-point, not per-boundary mocks.} A baseline that stubs every
boundary with its recorded return never re-enters the tool, so its verdict
cannot depend on the tool's code: the mock returns the recorded unsafe result
whether or not a gate exists. On all \Nincidents{} incidents it fails the
unguarded code, the guarded fix, and the benign edit alike
(Table~\ref{tab:results}). Chronicle's plan stubs the first model crossing and
runs the tool live, so the gated tool executes against the recorded arguments.
On the trade-notional incident the unguarded tool sells 1{,}000 shares
($\approx\$190$k) for a $\approx\$1$k request; the gated tool blocks it; the
cut-point test fails on the unguarded code and passes on the gated and benign
rewordings. The same pattern runs for all \Nincidents{} incidents.

\paragraph{Task and metrics.} A method receives a recorded trace and a code
version and returns fail or pass. We report on the curated suite:
(i)~fault-detection rate (unguarded fail, gated and benign pass);
(ii)~determinism of full replay (0 divergences over \determReps{} repetitions);
(iii)~provider calls and dollar cost under full replay; (iv)~recording overhead
per crossing relative to a typical model call; (v)~wall-clock time for one
full-stub and one cut-point pass over the suite; (vi)~boundary-annotation
coverage on the released agents (\boundarySites{} sites, three per incident, all
annotated explicitly); and (vii)~the share of mutants of each guarded tool that
a test kills. Evaluation is scoped to these \Nincidents{} incidents; we do not
claim reproduction rates for traces outside the suite.

\section{Experiments}
\label{sec:experiments}
\paragraph{Setup.} A single harness runs each item under recording and replay
and, for each code version, the cut-point test. Determinism is measured by
re-running the full-stub replay suite \determReps{} times. Separately, we time
\liveN{} calls per arm to \liveModel{}, an open-weight model served locally
(4-bit, Ollama on a laptop CPU), through a model boundary with recording on
versus off (interleaved pairs in random order) to place instrumentation cost
against real inference latency and its variance.

\paragraph{Recording overhead.} In-memory recording adds a median
\recOverheadUs\,\textmu s per crossing, or \recOverheadPct\% of an assumed
\modelLatencyMs~ms model call, and the store grows by at most \storeKB~KB per
crossing (Table~\ref{tab:results}). Against real \liveModel{} calls, mean
latency is \liveOnMean~ms (s.d.\ \liveOnStd~ms) with recording on and
\liveOffMean~ms (s.d.\ \liveOffStd~ms) with it off; the difference of means is
\liveDiff~ms (95\% CI \liveCIlo{} to \liveCIhi~ms), indistinguishable from zero.
Chronicle's measured cost of \recOverheadUs\,\textmu s per crossing is four
orders of magnitude below this run-to-run variation, so recording is not a
detectable source of latency.

\paragraph{Determinism and replay cost.} Full replay is bit-stable: across
\determReps{} repetitions it reproduces each recorded run identically, with 0
divergences and 0 live boundary crossings. Because the benchmark's model
boundaries are simulated, this validates the replay mechanism rather than
reproduction under a nondeterministic provider (see Limitations). A cut-point
test inherits this stability for its stubbed crossings, while its live crossings
re-execute the annotated functions. Full replay issues 0 of the \liveCalls{}
agent model-boundary crossings in the suite, so the provider cost of 1{,}000
full-suite replays is \$\replayCostK{}; one full-stub pass over the suite
completes in \replaySuiteMs~ms and one cut-point pass (tool and final model
crossing live) in \cutpointSuiteMs~ms (Table~\ref{tab:results}). A suite of
recorded incidents thus runs on every commit at no model cost.

\paragraph{Fault detection.} On the \Nincidents{} curated incidents, cut-point
tests give the expected fail\,/\,pass\,/\,pass on \faultCorrect{} of
\Nincidents{} (the unguarded code is caught; the gated fix and benign rewordings
pass) and tolerate \benignSurvived{} unrelated rewordings that leave the safety
invariant unchanged (Table~\ref{tab:results}). Because the incidents, guards,
and assertions were written together, these outcomes validate the mechanism on
curated incidents rather than measure fault detection in general.

\paragraph{Mutation study and full-stub baseline.} To test whether the
assertions catch faults beyond the known missing guard, we generate all
first-order mutants of each guarded tool with standard relational, logical, and
constant mutation operators (\mutantsTotal{} in total); a mutant is killed when
a test's verdict on it differs from its verdict on the unmutated fix. Cut-point
tests kill \mutantsKilledCP{}; the full-stub baseline kills none, since the tool
never runs under it. Of the \mutantsSurvived{} survivors, \mutantsUnkillable{}
cannot be killed by any test built from these recordings: they change code the
recorded input never reaches, or change the guard without changing its decision
on that input (e.g., $>$ to $\geq$ on a threshold the input exceeds). The other
\mutantsIgnoredField{} change output fields the assertion deliberately ignores,
such as the status label, while still blocking the action, so no surviving
mutant lets the recorded unsafe action through. Among mutants that change
behavior on the recorded input, cut-point tests thus kill \mutantsKilledCP{} of
\mutantsKillable{} (\mutantsKillableScore\%). The unkillable mutants reflect a
coverage limit of single-incident fixtures: each recording exercises one input.

\paragraph{Workflow.} A team records a production incident once, commits the
trace and a cut-point assertion to the repository, and the test then runs on
every commit with no model calls. When a boundary's contract changes so that a
stubbed name is crossed a different number of times, the count check flags the
fixture for re-recording rather than passing silently, which keeps the committed
test honest as the agent evolves.

\begin{table}[t]
\centering
\small
\setlength{\tabcolsep}{3pt}
\begin{tabular}{@{}lrr@{}}
\toprule
 & Cut-point & Full-stub \\
\midrule
\multicolumn{3}{@{}l}{\emph{Test outcomes}} \\
Fails unguarded code & \faultCorrect{}/\Nincidents{} & \Nincidents{}/\Nincidents{} \\
Passes fix + benign edit & \faultCorrect{}/\Nincidents{} & \fullstubSpec{}/\Nincidents{} \\
Rewordings tolerated & \benignSurvived{}/\benignSurvived{} & -- \\
Mutants killed & \mutantsKilledCP{}/\mutantsTotal{} & \mutantsKilledFS{}/\mutantsTotal{} \\
\midrule
\multicolumn{3}{@{}l}{\emph{Replay}} \\
Live model crossings & \Nincidents{}/\liveCalls{} & 0/\liveCalls{} \\
Divergences (\determReps{} runs) & -- & 0 \\
Suite pass (ms) & \cutpointSuiteMs & \replaySuiteMs \\
\midrule
\multicolumn{3}{@{}l}{\emph{Recording}} \\
Per crossing (\textmu s) & \multicolumn{2}{r}{\recOverheadUs} \\
\quad vs.\ \modelLatencyMs~ms call & \multicolumn{2}{r}{\recOverheadPct\%} \\
Store per crossing (KB) & \multicolumn{2}{r}{$\leq$\,\storeKB} \\
Model call, on / off (ms) & \multicolumn{2}{r}{\liveOnMean{} / \liveOffMean} \\
\bottomrule
\end{tabular}
\caption{Harness results on the released fixtures. Cut-point stubs the first
model crossing and runs the tool and the second model crossing live; full-stub
stubs every boundary. All \boundarySites{} boundary sites (three per incident)
are annotated explicitly with \texttt{@boundary}.}
\label{tab:results}
\end{table}

\section*{Limitations}
Chronicle does not capture streaming responses (recorded as their assembled
form) or concurrent parallel tool calls, and replay does not yet re-raise an
exception recorded at a stubbed boundary. An envelope captures a boundary's
interface, not its internal side effects, so a live cut-point on a destructive
tool should target a sandbox. Determinism here is partly by construction: the
released agents replace the model with a deterministic stub, so we do not
measure reproduction on a live nondeterministic provider. The benchmark is small
and self-constructed: six three-step incidents without the loops, retries, or
multi-agent routing that motivate this work. Count-preserving reorderings escape
the per-name check, and coverage of non-deterministic call sites is the user's
responsibility. The LLM-as-judge is implemented but not evaluated for
reliability \citep{norman2026reliability}.

\section*{Ethics Statement}
A recording copies prompts, agent state, and tool arguments, and may contain
secrets or personal data. Chronicle applies redaction at record time, before any
record is written or committed, preserving the structure that tests assert on
while removing sensitive values.

\bibliography{custom}

\appendix
\section{Additional Replay Diagrams}
\label{app:figures}
Figures~\ref{fig:fullreplay} and~\ref{fig:multilive} show the other two replay
modes, and Figure~\ref{fig:indexing} illustrates boundary indexing, expanding the
cut-point diagram in Figure~\ref{fig:cutpoint}.

\begin{figure}[ht]
\centering
\resizebox{\columnwidth}{!}{%
\begin{tikzpicture}[
  font=\scriptsize\ttfamily,
  box/.style={draw, rounded corners=2pt, align=center, minimum width=1.7cm,
              minimum height=1.05cm, inner sep=3pt, fill=stubfill},
  arr/.style={-{Latex}, thick},
]
\node[box] (a1) {agent[1]\\stub $\cdot$ recorded};
\node[box, right=0.55cm of a1] (t1) {place\_order[1]\\stub $\cdot$ recorded};
\node[box, right=0.55cm of t1] (a2) {agent[2]\\stub $\cdot$ recorded};
\draw[arr] (a1) -- (t1);
\draw[arr] (t1) -- (a2);
\end{tikzpicture}
}
\caption{Full replay: every boundary returns its recorded envelope (0 model calls;
bit-stable over \determReps{} reps).}
\label{fig:fullreplay}

\vspace{0.8em}
\resizebox{\columnwidth}{!}{%
\begin{tikzpicture}[
  font=\scriptsize\ttfamily,
  box/.style={draw, rounded corners=2pt, align=center, minimum width=1.7cm,
              minimum height=1.05cm, inner sep=3pt},
  stub/.style={box, fill=stubfill},
  live/.style={box, fill=livefill},
  arr/.style={-{Latex}, thick},
  note/.style={font=\scriptsize\sffamily},
]
\node[stub] (a1) {agent[1]\\stub model};
\node[live, right=0.5cm of a1] (t1) {place\_order[1]\\\textbf{live} tool};
\node[stub, right=0.5cm of t1] (a2) {agent[2]\\stub model};
\draw[arr] (a1) -- (t1);
\draw[arr] (t1) -- (a2);
\node[note, below=0.2cm of t1, text width=3.2cm, align=center]
  {live tool; models stubbed};
\end{tikzpicture}
}
\caption{Any subset may be live: tool live, both model crossings stubbed, so a
gate is tested without re-paying for model calls.}
\label{fig:multilive}

\vspace{0.8em}
\resizebox{\columnwidth}{!}{%
\begin{tikzpicture}[
  font=\scriptsize\ttfamily,
  box/.style={draw, rounded corners=2pt, align=center, minimum width=1.7cm,
              minimum height=1.1cm, inner sep=3pt, fill=stubfill},
  arr/.style={-{Latex}, thick},
  note/.style={font=\scriptsize\sffamily, align=left, text width=2.2cm},
]
\node[box] (a1) {agent[1]\\plan $\cdot$ tool\_call};
\node[box, right=0.55cm of a1] (t1) {place\_order[1]\\tool};
\node[box, right=0.55cm of t1] (a2) {agent[2]\\finalize $\cdot$ text};
\draw[arr] (a1) -- (t1);
\draw[arr] (t1) -- (a2);
\node[note, below=0.25cm of a1, text width=1.8cm] {1st \texttt{agent}};
\node[note, below=0.25cm of a2, text width=1.8cm] {2nd \texttt{agent}};
\end{tikzpicture}
}
\caption{Boundary indexing: the two model crossings are addressed as
\texttt{agent[1]} and \texttt{agent[2]}, by boundary name and occurrence rather
than by global step number.}
\label{fig:indexing}
\end{figure}

\end{document}